\documentclass{article}
\usepackage[preprint]{neurips_2023}

\usepackage[utf8]{inputenc} 
\usepackage[T1]{fontenc}    
\usepackage{hyperref}       
\usepackage{url}            
\usepackage{booktabs}       
\usepackage{amsfonts}       
\usepackage{nicefrac}       
\usepackage{microtype}      
\usepackage{xcolor}         
\usepackage{multirow}
\usepackage{listings}

\usepackage{subcaption}
\usepackage{enumitem}
\usepackage{algorithm}
\usepackage{algorithmicx}
\usepackage{hyperref}
\usepackage[noend]{algpseudocode}
\usepackage{graphicx}
\usepackage{tikz-cd}
\usepackage{tcolorbox}
 \usepackage[frozencache,cachedir=mint]{minted}
 \usepackage{fancyvrb}

 \usepackage{etoolbox}
\makeatletter
\patchcmd{\@verbatim}
  {\verbatim@font}
  {\verbatim@font\small}
  {}{}
\makeatother

\hypersetup{
   breaklinks=true,   
   colorlinks=true,   
}

\title{Scope is all you need:\\Transforming LLMs for HPC Code}

\author{%
  Tal Kadosh \\
  Ben-Gurion University, IAEC \\
  Israel \\
  \texttt{talkad@post.bgu.ac.il} \\
  \And
  Niranjan Hasabnis \\
  Intel Labs \\
  United States \\
  \texttt{niranjan.hasabnis@intel.com} \\
  \And
  Vy A. Vo \\
  Intel Labs \\
  United States \\
  \texttt{vy.vo@intel.com} \\
  \And
  Nadav Schneider \\
  Ben-Gurion University, IAEC \\
  Israel \\
  \texttt{nadavsch@post.bgu.ac.il} \\
  \And
  Neva Krien \\
  Independent Researcher \\
  Israel \\
  \texttt{nevo.krien@gmail.com} \\
  \And
  Abdul Wasay \\
  Intel Labs \\
  United States \\
  \texttt{abdul.wasay@intel.com} \\
  \And
  Nesreen Ahmed \\
  Intel Labs \\
  United States \\
  \texttt{nesreen.k.ahmed@intel.com} \\
  \And
  Ted Willke \\
  Intel Labs \\
  United States \\
  \texttt{ted.willke@intel.com} \\
  \And
  Guy Tamir \\
  Intel \\
  United States \\
  \texttt{guy.tamir@intel.com} \\
  \And
  Yuval Pinter \\
  Ben-Gurion University \\
  Israel \\
  \texttt{pintery@bgu.ac.il} \\
  \And
  Timothy Mattson \\
  Intel Labs \\
  United States \\
  \texttt{tim@timmattson.com} \\
  \And
  Gal Oren \\
  Technion, NRCN \\
  Israel \\
  \texttt{galoren@cs.technion.ac.il}
}

\newcommand{\tokom}[0]{\textsc{Tokompiler}}
\newcommand{\comp}[0]{\textsc{CompCoder}}

\begin{document}

\maketitle

\vspace{-0.6cm}
\begin{abstract}
\vspace{-0.3cm}
With easier access to powerful compute resources, there is a growing trend in the field of AI for software development to develop larger and larger language models (LLMs) to address a variety of programming tasks. Even LLMs applied to tasks from the high-performance computing (HPC) domain are huge in size (e.g., billions of parameters) and demand expensive compute resources for training. We found this design choice confusing --- \emph{why do we need large LLMs trained on natural languages and programming languages unrelated to HPC for HPC-specific tasks?}

In this line of work, we aim to question design choices made by existing LLMs by developing smaller LLMs for specific domains --- we call them \emph{domain-specific LLMs}. Specifically, we start off with HPC as a domain and propose a novel tokenizer named \tokom{}, designed specifically for preprocessing code in HPC and compilation-centric tasks. \tokom{} leverages knowledge of language primitives to generate language-oriented tokens, providing context-aware understanding of code structure while avoiding human semantics attributed to them.

We applied \tokom{} to pre-train a state-of-the-art model, \comp{} (based on PolyCoder), for a Fortran, C, and C++ code corpus mined from GitHub. We evaluate the performance of these models against a conventional multilingual code LLM. Results demonstrate that \tokom{} significantly enhances code completion accuracy and semantic understanding compared to Byte-Pair Encoding (BPE) tokenizer in normalized-perplexity tests, down to $\sim$1.6 perplexity score. Our domain-specific dataset and tokenizer outperforms multilingual pre-trained models.

This research opens avenues for further advancements in domain-specific LLMs, catering to the unique demands of HPC and compilation tasks. The sources of this work are available at our GitHub
\textcolor{blue}{\href{https://github.com/Scientific-Computing-Lab-NRCN/Tokompiler}{Tokompiler}} repository.
\end{abstract}

\section{Introduction}
Recent breakthroughs in the field of AI have led significant attention to language models (LMs) due to their remarkable capabilities in natural language processing~\citep{min2021recent} and understanding. 
Large language models (LLMs) ~\citep{zhao2023survey}, particularly exemplified by models such as GPT-3~\citep{floridi2020gpt} and its successors~\citep{bubeck2023sparks}, have demonstrated the potential to grasp intricate linguistic structure and context, sparking exploration of their applicability beyond natural language tasks.
In parallel, the field of high-performance computing (HPC) has been tackling increasingly complex and data-intensive problems~\citep{reed2022reinventing}. The field of HPC has experienced notable advancements in hardware, software, and algorithms, resulting in substantial improvements in computational performance and efficiency~\citep{dongarra2022hpc, reed2023hpc}. Combining the two trends, the integration of LLMs into HPC workflows has emerged as a compelling avenue for innovation~\citep{chen2023lm4hpc}. For instance, several recent efforts have applied LLMs for automatically inserting OpenMP pragmas/MPI functions in code~\citep{chen2023learning, harel2023learning, kadosh2023advising, nichols2023modeling, schneider2023mpi, shen2023multigraph}, overcoming limitations of static tools~\citep{harel2020source, milewicz2021negative, mosseri2020compar, prema2017identifying, prema2019study}.

While existing LLMs have shown great promise on HPC-related tasks, there are a number of challenges with the current setup. For instance, several of the existing LLMs that are applied to HPC tasks are pre-trained on natural languages and then fine-tuned on code corpus of several programming languages\footnote{This is because some of the code-related tasks, such as code summarization, demand a semantic understanding of natural languages.}. This design, however, leads to large models with billions of parameters that demand expensive compute resources for training and even inference. For instance, HPC-Coder~\citep{nichols2023modeling}, a recently-introduced LLM for HPC tasks, is obtained by finetuning PolyCoder~\citep{xu2022systematic} on an HPC dataset, while PolyCoder itself is a code LLM (not specific to HPC) that is trained on a 249GB code corpus of 12 programming languages. Such setups seemed counter-intuitive to us --- \emph{Is it not enough to train an LLM on HPC-specific languages only? In other words, why do we need an LLM trained on Java or Python (i.e., PolyCoder) for HPC-specific tasks?} More importantly, we believe that such \emph{domain-specific LLMs} would be computationally as well as financially efficient to train.


In this line of work, we hypothesize that domain-specific LLMs (e.g., smaller LMs that are designed and trained specifically on HPC datasets) would perform better than existing LLMs. Towards that end, we plan to revisit and evaluate each and every design decision made by existing LLMs with an eye toward solving HPC tasks. In this preliminary paper, we present our study on the first such design decision. Specifically, we propose a novel tokenization method, called \tokom{}, that focuses on code structure and language primitives by ensuring that all tokens are meaningful for specific code-language comprehension. Our method reduces the total amount of tokens compared to the standard Byte-Pair Encoding (BPE) method, allowing us to reduce the model size and improve on training time. This suggests that \tokom{} can also help in addressing the prohibitive computational and memory demands that continue to pose a significant obstacle to the practical adoption of LLMs for HPC languages~\citep{li2023starcoder, xu2022systematic}.

\section{\tokom{}: Code Tokenizer for HPC Code}
Tokenizing code for LLMs necessitates specialized techniques to accommodate programming language syntax.\footnote{These approaches include utilizing BPE and subword tokenization akin to natural language~\citep{sennrich2015neural}, employing syntax-aware tokenization to identify language-specific elements like keywords and identifiers~\citep{zheng2022code}, constructing tokens based on the Abstract Syntax Tree (AST) of code to capture structural information~\citep{xu2022survey}, implementing language-specific lexers to break code into tokens following grammar rules~\citep{bui2023codetf}, considering character-level tokenization to preserve character integrity~\citep{kc2023neural}, or even developing custom tokenization methods tailored to unique syntax rules, and leveraging dedicated code tokenization libraries.} LLMs geared towards code comprehension, such as GPT-3.5-Turbo for code,
likely combine several techniques, prioritizing syntax-aware tokenization to effectively process and generate code snippets in various programming languages and tasks \citep{ye2023comprehensive}. The chosen tokenization strategy hinges on factors such as the target programming language, intended code-related task, and codebase complexity.

\begin{figure}[t]
    \centering
\noindent
\begin{minipage}{0.3\linewidth}
\begin{verbatim}
// Source code:
int main() {
  int r[2800 + 1];
}
\end{verbatim}
\end{minipage}%
\begin{tikzpicture}
  \draw[->] (0,0) -- (0.4,0);
\end{tikzpicture}%
\hspace{0.1cm}
\begin{minipage}{0.3\linewidth}
\begin{verbatim}
// Tokompiler:
int func_252() {
 int arr_88[num_34 +
            num_842];
}
\end{verbatim}
\end{minipage}%
\begin{tikzpicture}
  \draw[->] (0,0) -- (0.4,0);
\end{tikzpicture}%
\hspace{0.1cm}
\begin{minipage}{0.3\linewidth}
\begin{verbatim}
// Lexicalized tokens:
["int", "func", "252", 
"(", ")", "{", "int",
... (tokens continue)
\end{verbatim}
\end{minipage}
    \caption{\tokom{} pipeline overview: Given a source code, the code turns into a semantic-less version using AST knowledge, and eventually, the lexicalized tokens are fed into \comp{}.}
    \label{fig:orderoftokom}
\end{figure}

In contrast to common tokenizers, we propose \tokom{}, a tokenization approach 
specifically targeting high-performance computing and compilation tasks. The \tokom{} tokenization process (demonstrated in \autoref{fig:orderoftokom}) involves generating an \textit{anonymized} version of the original code by replacing variable names, numbers, and strings; parsing this anonymized code to create an Abstract Syntax Tree (AST); updating the AST to reflect anonymization changes and maintaining a \textit{one-to-one change dictionary}; converting the modified AST back into code while discarding extraneous details; splitting multi-part tokens like variable names for improved understanding; and \textit{attaching random numbers} from a predefined range to recurring tokens to reduce reliance on specific replacements. 
In detail, the \tokom{} tokenization process for any language goes as follows:
\begin{enumerate}
    \item \textbf{Generate Replaced Code}: Create a version of the original code with anonymized variable names, numbers, and strings. The intuition behind this step is to eliminate misleading semantics such as variable \texttt{i} being an index variable of a \texttt{for} loop in C language.
    \item \textbf{AST Generation}: Parse the anonymized code using TreeSitter\footnote{\url{https://github.com/tree-sitter/tree-sitter}} or any suitable parser to generate an AST.
    \item \textbf{Recreate AST Changes}: Update the AST to reflect the changes made during anonymization. Keep a dictionary of all changes per file/function to facilitate restoring the semantics later.
    \item \textbf{AST to Code-Tokenize}: Transform the updated AST back into code, thus eliminating any comments, new lines, and READMEs that may interfere with anonymization. Although, removal of natural language (NL) from code may hamper ability of the model to solve NL tasks, the compilation tasks do not require NL understanding. More importantly, this code-tokenized version will have a much smaller number of tokens.
    \item \textbf{Token Splitting}: Split multi-part tokens (e.g., \texttt{var\_1} to [\texttt{var}, \texttt{1}]) to ensure that the model comprehends variable names as a combination of a type and a unique identifier.
    \item \textbf{Random Number Attachment}: For recurrent tokens (e.g., \texttt{var\_1} or \texttt{num\_2}), use statistics to attach random numbers from a predefined range (e.g., 1 to 1000) during tokenization. The attached numbers are randomly chosen without any relation to the type or order of the replaced tokens or the file/function length. This step also eliminates misleading semantics. For instance, if variable \texttt{i} is getting replaced with \texttt{var\_1} consistently, then the model may learn that \texttt{var\_1} is an index variable of \texttt{for} loops.
\end{enumerate}

\begin{figure}[t]
\centering
  \includegraphics[width=1\textwidth]{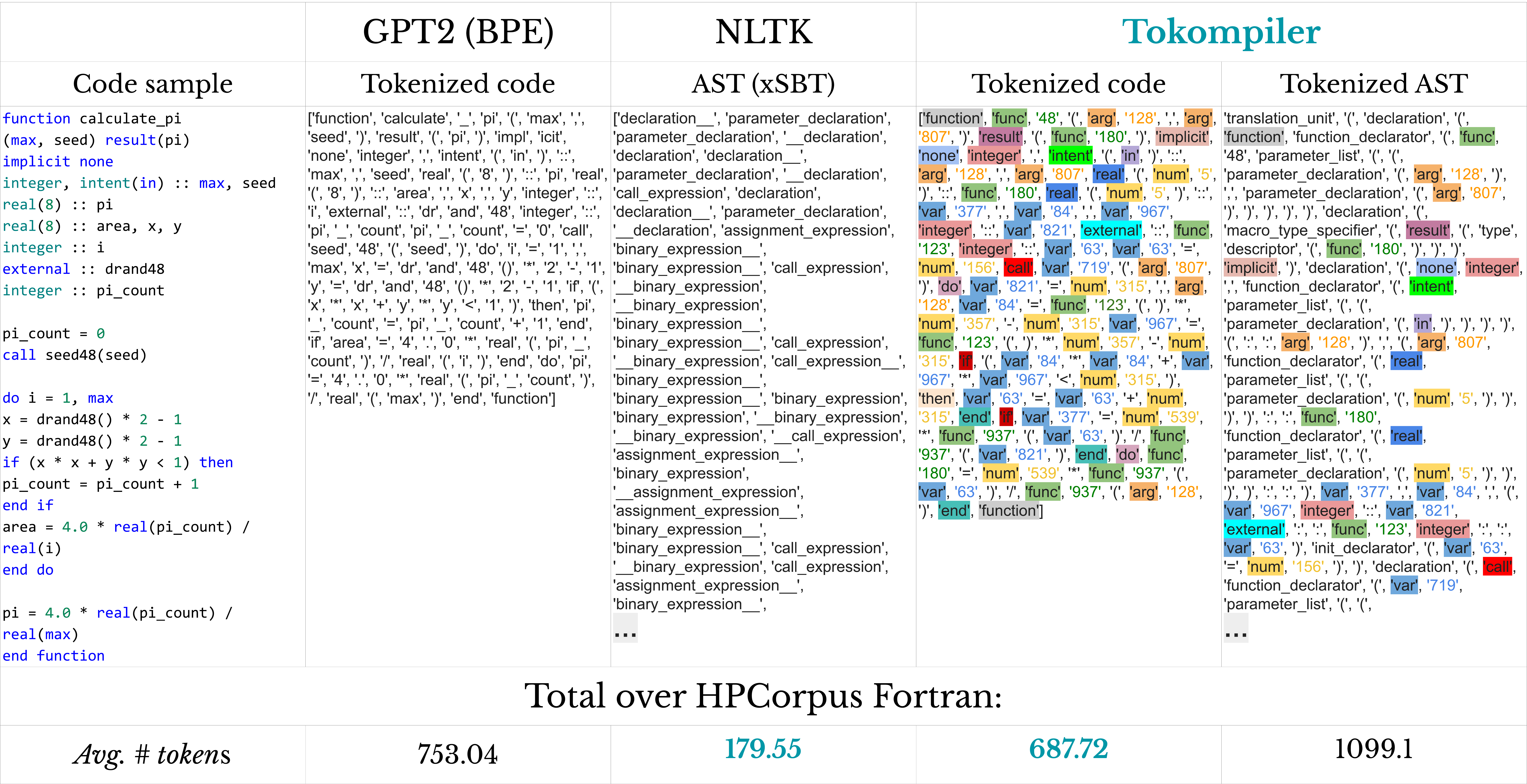}
  \caption{A code example with different tokenizers, and the resulting token count. Comparing the common byte-pair encoding (BPE) with our proposed \tokom{} showcases the dramatic decrease in the number of tokens while turning the source code into a machine-informed, rather than human-centered, representation.}
  \label{fig:tokompiler}
  \vspace{-0.2cm}
\end{figure}

\autoref{fig:tokompiler} demonstrates the difference between tokenization with GPT2-BPE, NLTK, and \tokom{} for code and AST, specifically showing the dramatic decrease in the total number of tokens for HPCorpus-Fortran~\citep{kadosh2023quantifying} (tokenized vocabulary is much smaller, 1177 against 50K) while enriching the tokenized source code with relevant information from the tokenized AST.

We hypothesize that \tokom{} enables language models to better understand code structure and context without memorizing specific misleading human semantics (such as variable names and function headers). Replacing those semantics with structure-aware representations (using AST data) leads to improved generalization and more accurate code completions. The method's ability to restore semantics back to the user using a dictionary of changes maintains the interpretability and usability of the model's outputs.


\section{\comp{}: pre-training a code LM on HPC code}

We evaluate \tokom{} by integrating it with a state-of-the-art code model, PolyCoder~\citep{xu2022systematic}. PolyCoder is a decoder-only transformer (like all GPT~\citep{floridi2020gpt} models) trained on a large multilingual corpus of source code. 
We henceforth name the integrated PolyCoder+\tokom{} model, \comp{}. The original PolyCoder model uses a GPT-style tokenizer (BPE~\citep{sennrich2015neural}) on a random 5\% subset of their dataset. Unlike BPE, \tokom{} has OOV tokens; however, their proportion is very small (0.08\%).

\textbf{\textit{Dataset.}}  We pre-trained both the PolyCoder and \comp{} models from scratch on (1) the Fortran dataset and on (2) the combined C and C++ dataset of HPCorpus~\citep{hpcorpus} (\autoref{tab:hpcorpus}). HPCorpus is a collection of many of the Fortran, C, and C++ programs on GitHub (up to May 2023). Out of the original files, we extracted code samples with the following restrictions: Each code sample is pre-processed such that no natural language is included; only structured blocks from de-duplicated files were included; and only code greater than 100 tokens and less than 1MB (as done in original PolyCoder training) was included.

\textbf{\textit{Hardware.}} We trained both original PolyCoder and \comp{} on 4 A40 48GB GPUs.

\textbf{\textit{Experimental setup.}} Although replacing the BPE tokenizer with \tokom{} reduces the embedding layer size, we matched all other model parameter sizes (e.g. number of heads, layers, hidden dimension, etc.) between PolyCoder and \comp{} to enable a fair comparison between the two. From the original PolyCoder architectures, we experimented with the small (162M parameter) model for Fortran and the large (2.8B parameter) model for the C and C++ data, following the original architecture configurations~\citep{xu2022systematic}. Since previous work~\citep{hoffman2022chinchilla} has indicated that model sizes should be scaled to the token count of the training data, we also ran experiments that reduced the number of layers in the models to approximate a 20:1 token:parameter ratio for assuming the use of \tokom{}. For Fortran, this meant reducing the number of layers from 12 to 8, and for C/C++ reducing 32 layers to 8 layers.

All pre-training experiments were trained with the Adam optimizer, and followed a learning rate schedule with a linear warmup for the first 10\% of steps and a cosine decay over the remaining steps. For the Fortran experiments, we trained the models for 60K steps at a learning rate of 0.00025. We also applied regularization, applying a 0.01 weight decay, a 0.1 dropout rate. For the C/C++ experiments, we trained the models for 160K steps at a learning rate of 0.00008. We did not find regularization necessary for these experiments.

Finally, we performed one fine-tuning experiment using the 2.8B PolyCoder model pre-trained on the multilingual corpus from ~\citep{xu2022systematic}. Since the original corpus did not contain Fortran, we only fine-tuned it on the C and C++ training set of HPCorpus for 35K steps at a 0.00016 learning rate, following the same schedule as above.

\begin{table}
\centering
\begin{footnotesize}
\begin{tabular}{|c||r|r|r|r|}
\hline
                 & \textbf{Repos} & \textbf{Size(GB)} & \textbf{Files (\#)} & \textbf{Functions (\#)} \\ \hline
Fortran & 3,683          & 0.68              & 138,552             & 359,272                 \\ \hline
C       & 144,522        & 46.23             & 4,552,736           & 87,817,591              \\ \hline
C++     & 150,481        & 26.16             & 4,735,196           & 68,233,984              \\ \hline
\end{tabular}
\end{footnotesize}
\vspace{0.2cm}

\caption{Statistics on the HPCorpus dataset: a total of $\sim$300K repos, $\sim$70 GB, $\sim$9M files, and $\sim$155M functions across Fortran, C and C++ code from GitHub.
}
\label{tab:hpcorpus}
  \vspace{-0.2cm}
\end{table}


\textbf{\textit{Results.}} We measured the normalized perplexity of both PolyCoder and \comp{} and reported the best values for each model type across experiments.\footnote{Inspired by LM-PPL \citep{lmppl} library as described in~\citep{xu2022systematic}. A fork of \citep{lmppl} is available in our public fork.} 
While the BPE and \tokom{} tokenizers have different vocabulary, we are confident that comparing them is meaningful thanks to the work in \citep{erdmann2019little}.
LM perplexity is an important measure for downstream HPC tasks, such as OpenMP pragma generation.\footnote{In this task, as exemplified in \citep{nichols2023modeling}, a subset of OpenMP programs has been created, in which the pragma is moved to the end of the structured block, and the task is to generate the pragma based on the structured block itself. This method can also be applied to the generation of other parallel APIs, such as the incremental insertion of MPI~\citep{gabriel2004open} functions for domain decomposition~\citep{schneider2023mpi}, or for other accelerated computing APIs such as SYCL~\citep{reyes2016sycl} or OpenACC~\citep{farber2016parallel}.}

\begin{table}
\centering
\begin{footnotesize}
\begin{tabular}{|c||c|c||c|c|}
\hline
 &
 \multicolumn{2}{c||}{{Fortran}} &
    \multicolumn{2}{c|}{{C \& C++}} \\ \hline

Model   & PolyCoder  & \comp{} & PolyCoder & \comp{} \\ \hline
Code (perplexity)  & 2.46 & \textbf{1.59} & 1.93  & \textbf{1.65} \\ \hline
Model Size & 162M & \textbf{59M} & 2.8B & \textbf{638M} \\ \hline
Time-to-train (mins) & 435 & \textbf{246} & 8300 & \textbf{2066} \\ \hline 
\end{tabular}
\end{footnotesize}
\vspace{0.2cm}
\caption{Performance of PolyCoder (BPE) vs. \comp{} (\tokom{}) pre-trained on HPCorpus.} 
\label{tab:perplexity}
  \vspace{-0.4cm}
\end{table}

We found that \comp{} fared better with the smaller model architectures, resulting in a significant reduction of training time (\autoref{tab:perplexity}). The results show that not only did the new given form of data (stripped of natural language and using structured blocks) help the models in improving the results, but the usage of \tokom{} improved language modeling performance by 35\% on Fortran and 15\% on C and C++. By comparison, fine-tuning the 2.8B PolyCoder model that had been pre-trained on the multilingual corpus with BPE fared worse than either model pre-trained on the restricted, domain-specific HPCorpus, achieving a 2.20 test perplexity on the same test set.

We stress that the results obtained using the \tokom{} are even more impressive than just an improvement upon the previously measured perplexity. Since we removed any human semantics, we also proved -- for the first time -- that it is possible to successfully pre-train a language model to understand the actual design patterns of code and prove that such a model understands the code behavior. 

\section{Future work}
In the near future, we intend to apply \tokom{} to C and C++ corpora and integrate more code representations, such as the data-flow graph (DFG) and the intermediate representation (IR)~\citep{grossman2023compile}, to enhance model understanding as shown in closely related works ~\citep{szafraniec2022code, guo2020graphcodebert}. We also intend to fine-tune those pre-trained models for HPC downstream tasks, such as OpenMP pragma generation~\citep{nichols2023modeling,kadosh2023advising} and MPI domain decomposition distribution~\citep{nichols2023modeling, schneider2023mpi}. In general, given the differences between general programming tasks and HPC-specific programming tasks, our research vision is to systematically analyze each and every element of existing LLMs (model architecture, dataset, etc.) and redesign them as needed for HPC-specific tasks.

\begin{ack}
This research was supported by the Israeli Council for Higher Education (CHE) via the Data Science Research Center, Ben-Gurion University of the Negev, Israel; Intel Corporation (oneAPI CoE program); and the Lynn and William Frankel Center for Computer Science. Computational support was provided by HPE HPC \& AI Cloud \citep{breckenridge}, Intel Developer Cloud~\citep{intel-cloud}, and the NegevHPC project~\citep{negevhpc}.
\end{ack}

\bibliographystyle{plainnat}
\bibliography{bibliography}

\end{document}